\documentclass[11pt,a4paper]{article}

\usepackage[utf8]{inputenc}
\usepackage[T1]{fontenc}
\usepackage{amsmath}
\usepackage{amssymb}
\usepackage{graphicx}
\usepackage{hyperref}
\usepackage{geometry}
\usepackage{setspace}
\usepackage{authblk}
\usepackage{booktabs}
\usepackage{caption}
\usepackage{subcaption}
\usepackage{xcolor}
\usepackage[numbers,sort&compress]{natbib}
\usepackage{titlesec}
\usepackage{enumitem}
\usepackage{float}
\usepackage{array}
\usepackage{multirow}
\usepackage{url}
\usepackage{lineno}
\usepackage{tabularx}
\usepackage{longtable}
\usepackage{makecell}

\hypersetup{
    colorlinks=true,
    linkcolor=blue,
    citecolor=blue,
    urlcolor=blue
}

\titleformat{\section}{\large\bfseries}{\thesection.}{0.5em}{}
\titleformat{\subsection}{\normalsize\bfseries}{\thesubsection.}{0.5em}{}
\titleformat{\subsubsection}{\normalsize\itshape}{\thesubsubsection.}{0.5em}{}

\title{%
\rule{\textwidth}{2pt}\\
{\LARGE\textbf{Small Language Models for Privacy-Preserving Clinical
Information Extraction in Low-Resource Languages}}\\[0.5ex]
\rule{\textwidth}{2pt}
}

\author{%
    \large
    Mohammadreza Ghaffarzadeh-Esfahani$^{1,\dagger}$,\;
    Nahid Yousefian$^{1}$,\;
    Ebrahim Heidari-Farsani$^{1}$,\;
    Ali Akbar Omidvarian$^{1}$,\
    Sepehr Ghahraei$^{1}$,\;
    Atena Farangi$^{1}$,\;
    AmirBahador Boroumand$^{2,\dagger}$\\
    \small\itshape
    $^{1}$Student Research Committee, Isfahan University of Medical Sciences, Isfahan, Iran\\
    $^{2}$Department of Emergency Medicine, Isfahan University of Medical Sciences, Isfahan, Iran\\
    \small\normalfont
    $^{\dagger}$Correspondence:\\
    Mohammadreza Ghaffarzadeh-Esfahani:
    \href{mailto:mreghafarzadeh@gmail.com}{mreghafarzadeh@gmail.com},\;
    Tel/Fax: +98-3136700479,\;
    ORCID: 0009-0009-9322-5471\\
    AmirBahador Boroumand:
    \href{mailto:ab.boroumand@med.mui.ac.ir}{ab.boroumand@med.mui.ac.ir},\;
    Tel/Fax: +98-36688597,\;
    ORCID: 0000-0001-5055-6881
}
\date{\rule{\textwidth}{0.4pt}}

\begin{document}

\maketitle

\begin{abstract}
Extracting clinical information from medical transcripts in low-resource languages remains a significant challenge in
healthcare natural language processing (NLP). This study evaluates a
two-step pipeline combining Aya-expanse-8B as a Persian-to-English
translation model with five open-source small language models (SLMs)
--- Qwen2.5-7B-Instruct, Llama-3.1-8B-Instruct, Llama-3.2-3B-Instruct,
Qwen2.5-1.5B-Instruct, and Gemma-3-1B-it --- for binary extraction of
13 clinical features from 1,221 anonymized Persian transcripts collected
at a cancer palliative care call center. Using a few-shot prompting
strategy without fine-tuning, models were assessed on macro-averaged
F1-score, Matthews Correlation Coefficient (MCC), sensitivity, and
specificity to account for class imbalance. Qwen2.5-7B-Instruct achieved
the highest overall performance (median macro-F1: 0.899; MCC: 0.797),
while Gemma-3-1B-it showed the weakest results. Larger models (7B--8B
parameters) consistently outperformed smaller counterparts in sensitivity
and MCC. A bilingual analysis of Aya-expanse-8B revealed that
translating Persian transcripts to English improved sensitivity, reduced
missing outputs, and boosted metrics robust to class imbalance, though at the cost of
slightly lower specificity and precision. Feature-level results showed
reliable extraction of physiological symptoms across most models, whereas
psychological complaints, administrative requests, and complex somatic
features remained challenging. These findings establish a practical,
privacy-preserving blueprint for deploying open-source SLMs in
multilingual clinical NLP settings with limited infrastructure and
annotation resources, and highlight the importance of jointly optimizing
model scale and input language strategy for sensitive healthcare
applications.

\vspace{0.5cm}
\end{abstract}

\newpage
\section{Introduction}

In the era of digital health, natural language processing (NLP) has
emerged as a transformative tool for data extraction from unstructured
clinical text~\cite{ref1}. This extraction is crucial in medicine, as
it converts raw descriptions into structured, quantifiable
information, facilitating advanced analytics like predictive modeling for
disease progression~\cite{ref2} and evidence-based decision support
systems that enhance patient outcomes and operational efficiency~\cite{ref3}.
Particularly in palliative oncology, where patients often report
multifaceted symptoms, automated information extraction from
patient-provider interactions can alleviate clinician workload, reduce
diagnostic delays, and optimize support, especially in
resource-constrained settings~\cite{ref4}. However, the practical
implementation of NLP solutions, especially across diverse languages and
settings, faces technical and methodological hurdles~\cite{ref5}.

Traditional NLP pipelines for medical entity recognition and data
extraction have relied on rule-based systems~\cite{ref6} or supervised
machine learning models trained on annotated corpora~\cite{ref7}, but
these approaches face challenges like training data scarcity and struggle
to understand linguistic nuances~\cite{ref8}. Recently, the advent of
large language models (LLMs) like GPT-5~\cite{ref9} has promised
capabilities for cross-lingual tasks~\cite{ref10}, yet their proprietary
nature, reliant on API access, raises critical data safety and security
concerns, risking patient privacy~\cite{ref11}. Meanwhile, their large
open-source counterparts demand substantial computational resources,
posing barriers to deployment in low-infrastructure
environments~\cite{ref12}. This led to a growing interest in small
language models (SLMs) with compact architectures (1--10 billion
parameters), fine-tuned for instruction-following, which provide a
compelling balance of efficiency, accessibility, and
performance~\cite{ref13,ref14}. Recent benchmarks, such as those on the
MedNLI and MIMIC-III datasets, demonstrate SLMs' viability for
English-centric clinical tasks, achieving high F1-scores for symptom
detection~\cite{ref15}. Yet, empirical evidence on their efficacy for medical extraction in low-resource languages, especially Persian, remains understudied. Significant gaps persist in understanding how translation artifacts, class imbalance, and language-specific prompting affect performance in real-world clinical settings.

In this study, we address these gaps by designing a two-step pipeline
containing a translator model, Aya-expanse-8B~\cite{ref16}, plus five
open-source SLMs, Qwen2.5-7B-Instruct~\cite{ref17},
Llama-3.1-8B-Instruct~\cite{ref18}, Llama-3.2-3B-Instruct~\cite{ref19},
Qwen2.5-1.5B-Instruct~\cite{ref20}, and Gemma-3-1B-it~\cite{ref21}, on
a novel dataset of 1,221 anonymized Persian transcripts from a cancer
palliative care call center. Leveraging a few-shot prompting
strategy~\cite{ref22} for binary extraction of 13 clinical features, we
assess model performance across metrics robust to imbalance, including
macro-averaged F1-score, Matthews Correlation Coefficient (MCC),
sensitivity, and specificity. We additionally conducted a bilingual
assessment of Aya-expanse-8B, contrasting direct Persian processing
against English-translated inputs to quantify translation's trade-offs.
Our findings reveal that while larger SLMs, such as Qwen2.5-7B-Instruct
and Llama-3.1-8B-Instruct, delivered superior performance for symptom
detection, the use of original Persian inputs in the Aya-expanse-8B
model notably enhanced the recognition of more subjective conditions,
such as psychological complaints. These insights not only benchmark
SLMs' potential for equitable, low-resource healthcare NLP but also
inform hybrid strategies that combine native-language inference with
translation safeguards to enhance robustness in sensitive domains.

\section{Results}

\begin{figure}[!ht]
\centering
\includegraphics[width=0.8\textwidth]{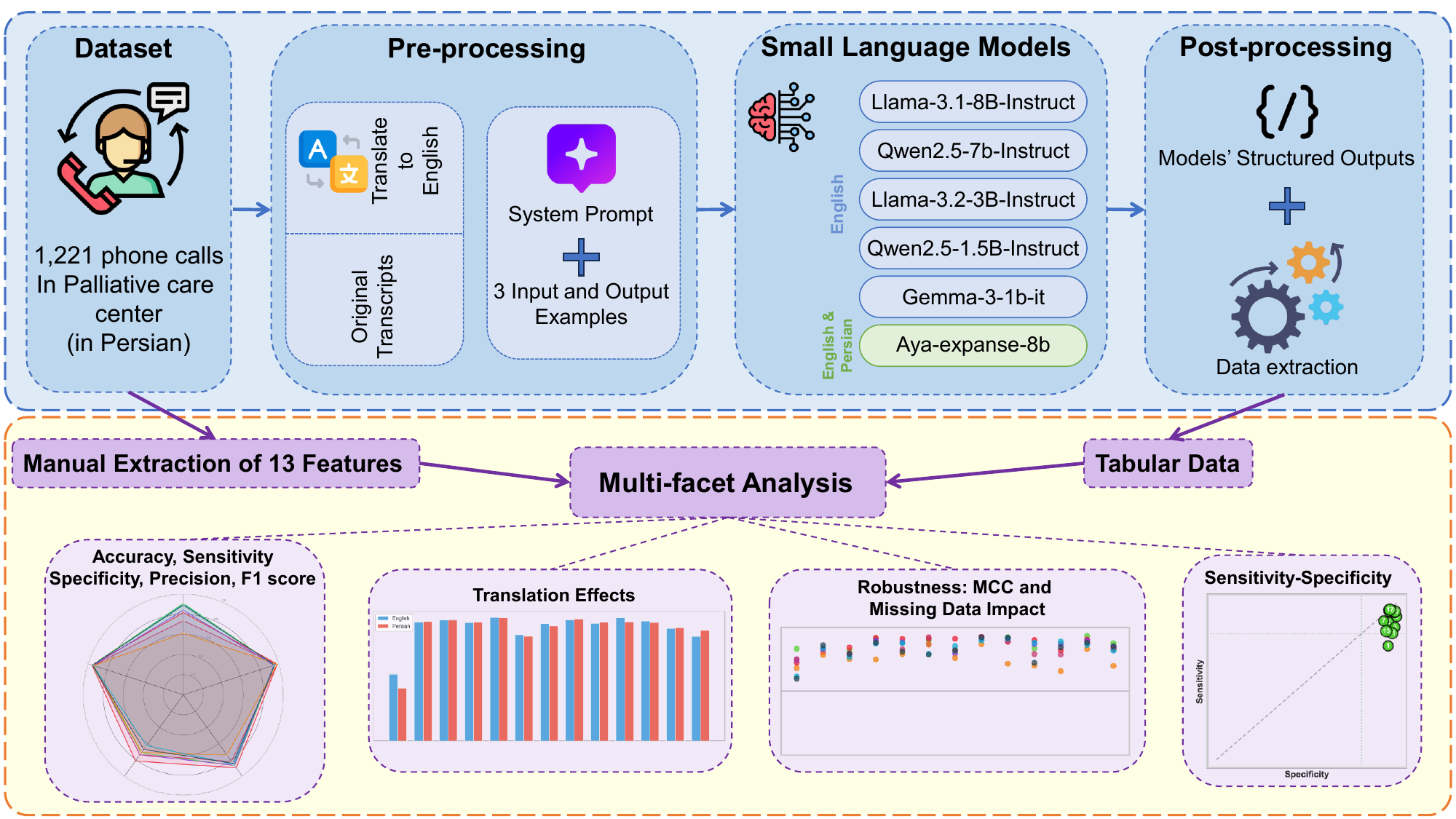} 
 \caption{\textbf{Schematic overview of the study.} The upper panel shows the dataset preprocessing, inference generation,
and postprocessing, starting from 1{,}221 Persian palliative care
phone-call transcripts, followed by translation into English, prompt
construction with input--output examples, and inference using multiple
small language models (SLMs). The models' structured outputs are then
post-processed to extract tabular data. The lower panel illustrates the
multi-facet analysis framework, comparing manual extraction of 13
reference features with model-derived features through performance
metrics (accuracy, sensitivity, specificity, precision, F1-score),
assessment of translation effects, crobustness analysis (Matthews correlation coefficient (MCC), missing values), and sensitivity--specificity trade-offs.}
\label{fig:1}
\end{figure}

Our evaluation of the proposed two-step pipeline, which combines the
Aya-expanse-8B translator model with five open-source SLMs, demonstrated
variability in the extraction of structured clinical information from
Persian palliative oncology call transcripts. Benchmarking against a
manually annotated dataset of 1,221 calls revealed that model capability
was strongly influenced by parameter count and input language, with no
single model dominating across all 13 clinical features. To provide a
comprehensive overview of the models' capabilities, the results are
presented across four key dimensions: the overall comparative performance
of models, their feature-specific analysis, a focused analysis on the
impact of language translation for the multilingual model, and an
evaluation of model robustness through imbalance-aware performance metrics and error analysis.
This multifaceted presentation reveals critical insights into models'
behavior, the influence of parameter scale and language, and the
practical implications of deploying such systems in sensitive domains
such as palliative care. A visual summary illustrating the key aspects
of the study is presented in Figure~\ref{fig:1}.

\subsection{Qwen2.5-7B-Instruct demonstrated the highest overall
performance among the evaluated models}

\begin{figure}[!ht]
\centering
\includegraphics[width=0.8\textwidth]{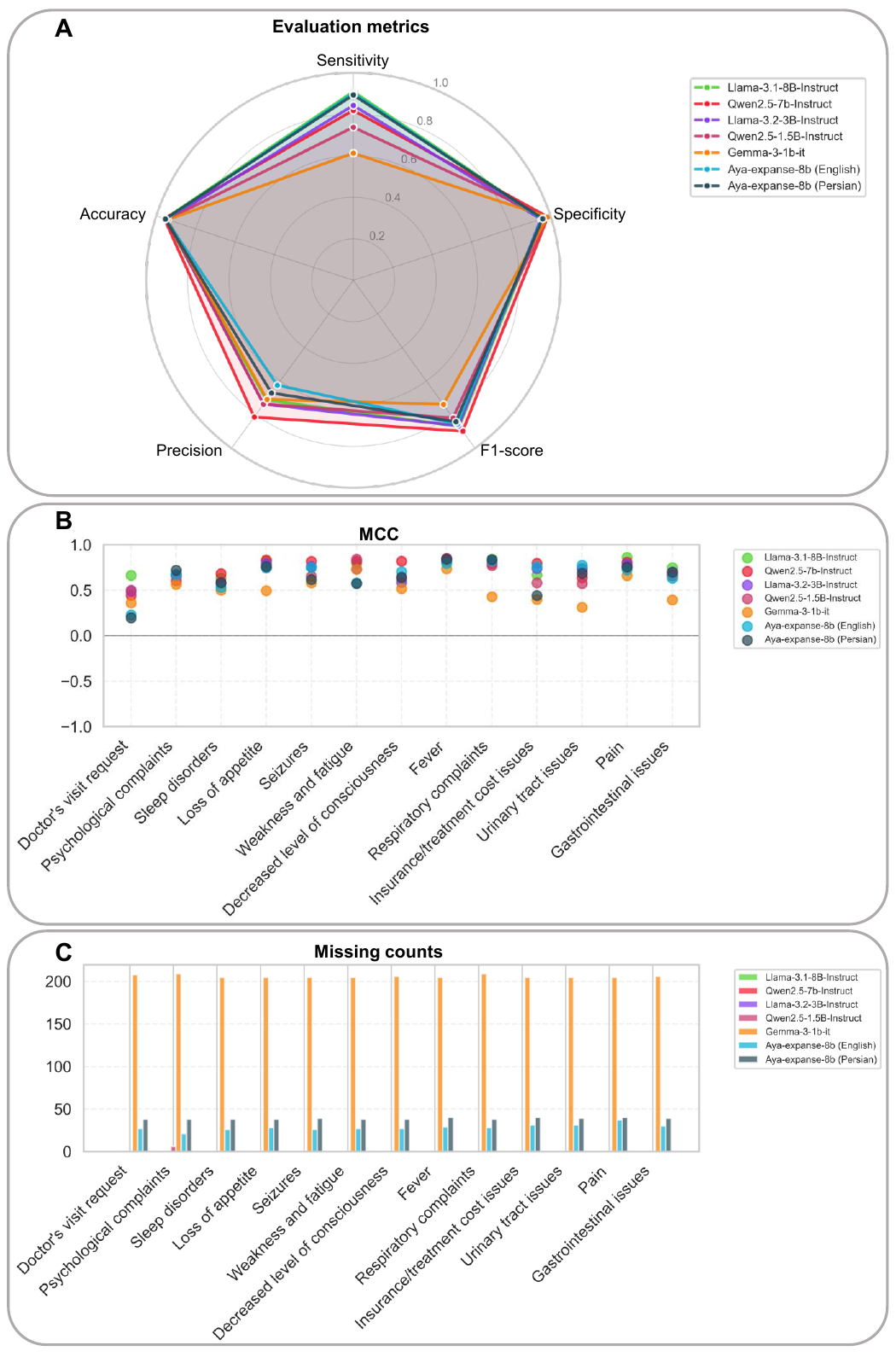} 
\caption{\textbf{Comparative performances of different models on validation metrics.} 
\textbf{(A)} The median value for 5 metrics of accuracy, sensitivity, specificity, macro-averaged F1 score, and precision among 13 extracted features compared to the manually-extracted ground truth. 
\textbf{(B)} Matthews Correlation Coefficient (MCC) values for each evaluated model across the 13 extracted clinical features, comparing model-generated outputs with the manually extracted ground-truth annotations. 
\textbf{(C)} Total number of missing counts for each model among different extracted features.}
\label{fig:2}
\end{figure}

Evaluation of the five SLMs demonstrated heterogeneous performance in extracting binary clinical features from the translated English transcripts. Overall, Qwen2.5-7B-Instruct exhibited the strongest
balanced performance, achieving the highest median specificity (0.987
[0.975, 0.992]), macro-averaged F1-score (0.899 [0.832, 0.908]),
precision (0.814 [0.759, 0.878]), and accuracy (0.96 [0.947, 0.984])
across all features. This model demonstrated robustness in handling
class imbalance (Figure~\ref{fig:2}A and Supplementary File 1).

In contrast, Gemma-3-1B-it showed the lowest median sensitivity (0.613
[0.294, 0.735]) and macro-averaged F1-score (0.74 [0.676, 0.79]),
reflecting challenges in detecting positive instances. Additionally, the
multilingual Aya-expanse-8B model performed comparably when using
different prompt variants (macro-averaged F1-score: English 0.855
[0.804, 0.87] and Persian 0.842 [0.753, 0.868]). Finally, larger models
like Llama-3.1-8B-Instruct and Qwen2.5-7B-Instruct generally
outperformed their smaller counterparts (e.g., Llama-3.2-3B-Instruct and
Qwen2.5-1.5B-Instruct) in sensitivity and overall accuracy, with median
accuracies exceeding 0.90 in most cases.

\subsection{Extraction performance was highly dependent on the type of
clinical feature}

To gain deeper insight into model performance, we conducted a
feature-wise analysis comparing model predictions with manually
extracted data. The results revealed substantial variability in how each model identified specific clinical features from
patients' phone calls. The following section presents a comparison of
macro-averaged F1 scores for each feature across all evaluated
models (Table~\ref{tab:1}). Other evaluation metrics, including accuracy,
precision, sensitivity, and specificity, are provided in Supplementary File 1.

\begin{table}[H]
\centering
\caption{\textbf{Macro-averaged F1 scores for the binary extraction of 13
clinical features from Persian palliative care transcripts across the
evaluated small language models.} Values are reported for
Llama-3.1-8B-Instruct, Qwen2.5-7B-Instruct, Llama-3.2-3B-Instruct,
Qwen2.5-1.5B-Instruct, Gemma-3-1B-it, and Aya-expanse-8B
(English-translated and direct Persian variants).}
\label{tab1}
\resizebox{\textwidth}{!}{%
\begin{tabular}{lccccc|cc}
\toprule
\multirow{2}{*}{\textbf{Feature}}
  & \textbf{Llama-3.1}   & \textbf{Qwen2.5}     & \textbf{Llama-3.2}   & \textbf{Qwen2.5}       & \textbf{Gemma-3}
  & \multicolumn{2}{c}{\textbf{Aya-expanse-8B}} \\
  & \textbf{-8B-Inst.}   & \textbf{-7B-Inst.}   & \textbf{-3B-Inst.}   & \textbf{-1.5B-Inst.}   & \textbf{-1B-it}
  & \textbf{English}     & \textbf{Persian} \\
\midrule
Doctor's visit request        & \textbf{0.832} & 0.678 & 0.723 & 0.733 & 0.633 & 0.493 & 0.411 \\
Psychological complaints      & 0.825 & 0.832 & 0.818 & 0.800 & 0.782 & 0.835 & \textbf{0.859} \\
Sleep disorders               & 0.804 & \textbf{0.838} & 0.766 & 0.788 & 0.740 & 0.739 & 0.753 \\
Loss of appetite              & 0.905 & \textbf{0.915} & 0.895 & 0.874 & 0.732 & 0.865 & 0.874 \\
Seizures                      & 0.869 & \textbf{0.908} & 0.880 & 0.819 & 0.790 & 0.869 & 0.798 \\
Weakness and fatigue          & 0.892 & 0.907 & 0.856 & \textbf{0.921} & 0.865 & 0.753 & 0.742 \\
Decreased level of consciousness & 0.792 & \textbf{0.909} & 0.769 & 0.802 & 0.748 & 0.843 & 0.804 \\
Fever                         & 0.907 & \textbf{0.924} & 0.907 & 0.921 & 0.867 & 0.891 & 0.918 \\
Respiratory complaints        & \textbf{0.918} & 0.891 & 0.908 & 0.883 & 0.655 & 0.908 & \textbf{0.918} \\
Insurance/treatment cost issues & 0.816 & \textbf{0.899} & 0.866 & 0.791 & 0.689 & 0.870 & 0.674 \\
Urinary tract issues          & 0.870 & 0.795 & 0.866 & 0.766 & 0.598 & \textbf{0.888} & 0.842 \\
Pain                          & \textbf{0.930} & 0.903 & 0.893 & 0.873 & 0.828 & 0.855 & 0.868 \\
Gastrointestinal issues       & \textbf{0.874} & 0.818 & 0.848 & 0.828 & 0.676 & 0.804 & 0.849 \\
\midrule
Median {[}IQR$_1$, IQR$_3${]}
  & \makecell{0.870\\{[}0.825,\\0.905{]}}
  & \makecell{\textbf{0.899}\\{[}0.832,\\0.908{]}}
  & \makecell{0.866\\{[}0.818,\\0.893{]}}
  & \makecell{0.819\\{[}0.791,\\0.874{]}}
  & \makecell{0.74\\{[}0.676,\\0.79{]}}
  & \makecell{0.855\\{[}0.804,\\0.87{]}}
  & \makecell{0.842\\{[}0.753,\\0.868{]}} \\
\bottomrule
\end{tabular}%
}
\label{tab:1}
\end{table}

\subsubsection{Physiological symptoms were reliably detected across most
models}

Several features were consistently identified with a high macro-averaged
F1 score ($>$0.85) by most models. Pain was the top-performing feature
overall, with Llama-3.1-8B-Instruct achieving the highest score (0.93).
Fever and respiratory complaints also showed robust performance, with
multiple models (Qwen2.5-7B-Instruct, Qwen2.5-1.5B-Instruct,
Aya-expanse-8B (Persian) for Fever; Llama-3.1-8B-Instruct,
Llama-3.2-3B-Instruct, Aya-expanse-8B (Persian) for respiratory
complaints) scoring above 0.91. Similarly, loss of appetite and seizures
were well-detected, particularly by the Qwen2.5-7B-Instruct model (0.915
and 0.908, respectively).

\subsubsection{Complex somatic and psychological features showed
inconsistent extraction}

Performance was more varied for features like weakness and fatigue,
decreased level of consciousness, and psychological complaints. While
some models excelled (for example, Qwen2.5-7B-Instruct on decreased
level of consciousness (0.909) and Aya-expanse-8B (Persian) on
psychological complaints (0.859)), others, particularly smaller models
like Gemma-3-1B-it, struggled more on these conceptually related
symptoms.

\subsubsection{Administrative requests and multifaceted complaints
remained difficult for all models}

Certain features proved challenging across the board. The doctor's visit
request was notably difficult, with the highest score (0.832 from
Llama-3.1-8B-Instruct) and the lowest score (0.410 from Aya-expanse-8B
(Persian)). Sleep disorders and insurance/treatment cost issues also
showed lower and more inconsistent scores. Features related to urinary tract and gastrointestinal issues revealed a performance gap with larger models like Llama-3.1-8B-Instruct handled them reasonably well (0.870 and 0.874, respectively), but the Gemma-3-1B-it model performed very poorly (0.598 and 0.676, respectively).

Ultimately, while all models reliably identified overt physiological
symptoms, their performance diverged significantly for administrative
requests, complex somatic complaints, and psychological expressions.
Notably, Aya-expanse-8B demonstrated language-dependent variability,
performing differently across English and Persian transcripts. This
suggests that the choice of input language may influence extraction
accuracy and completeness, highlighting the importance of evaluating
multilingual models in both original and translated contexts. The
following section examines how language translation affects
Aya-expanse-8B's performance across the same clinical features.

\subsection{Translating Persian transcripts to English enhanced
sensitivity and reduced missing outputs}

To assess the influence of translating Persian transcripts to English on
the performance of the multilingual Aya-expanse-8B model, we compared
its outputs when processing original Persian text directly (Persian
version) versus English-translated inputs (English version). This
analysis focused on the 13 binary features extracted from the 1,221
transcripts, using the evaluation metrics described in the methods
section. Overall, translation to English resulted in a modest improvement
in performance metrics, such as macro-averaged F1 score and MCC
(Figure~\ref{fig:2}B and Supplementary file 1). Notably, the English
version produced fewer missing predictions, suggesting enhanced output
completeness (Figure~\ref{fig:2}C and Supplementary File 1).

The English version achieved a macro-averaged F1 score of
0.855~[0.804--0.870] compared to 0.842~[0.753--0.868] for the Persian
version. Similarly, the MCC was higher in English (0.724~[0.634--0.757])
than in Persian (0.686~[0.584--0.757]), reflecting a stronger overall
correlation between predictions and ground truth labels. These gains were
driven by higher sensitivity in English (0.901~[0.842--0.911]) versus
Persian (0.893~[0.818--0.952]), which is particularly valuable for
detecting rare positive cases in this imbalanced dataset. However,
precision was lower in English (0.625~[0.597--0.671]) than in Persian
(0.672~[0.413--0.745]), leading to a higher rate of false positives and a
corresponding slight drop in accuracy (0.951~[0.905--0.973] vs.\
0.955~[0.912--0.963]) and specificity (0.955~[0.906--0.976] vs.\
0.960~[0.919--0.975]).

Feature-level analysis revealed that the English version outperformed
the Persian version in macro-averaged F1 for 6 features: doctor's visit
request (0.493 vs.\ 0.411), seizures (0.869 vs.\ 0.798), weakness and
fatigue (0.753 vs.\ 0.742), decreased level of consciousness (0.843 vs.\
0.804), insurance/treatment cost issues (0.870 vs.\ 0.674), and urinary
tract issues (0.888 vs.\ 0.842). Conversely, the Persian version excelled
in 7 features, including psychological complaints (0.859 vs.\ 0.835),
sleep disorders (0.753 vs.\ 0.739), loss of appetite (0.874 vs.\ 0.865),
fever (0.918 vs.\ 0.891), respiratory complaints (0.918 vs.\ 0.908),
pain (0.868 vs.\ 0.855), and gastrointestinal issues (0.849 vs.\ 0.804).

A key drawback of using Persian transcripts was the higher incidence of
missing values (38~[38, 39]) compared to 28~[27, 30] in English,
potentially arising from the model's challenges in adhering to the
structured output template in the native language. This incompleteness
disproportionately affected features such as fever and
insurance/treatment cost issues, where Persian missing outputs reached 40.

In summary, while direct processing of Persian transcripts yielded
marginally superior accuracy and specificity, translating to English
enhanced the model's robustness for imbalanced medical extraction tasks
by improving sensitivity, reducing missing outputs, and boosting balanced
metrics such as F1 and MCC. These findings underscore the potential
benefits of English-centric prompting for multilingual SLMs in
low-resource clinical settings.

\subsection{Larger models demonstrated superior performance under class
imbalance}

To assess the robustness of the evaluated SLMs in handling
class-imbalanced medical extraction tasks, we analyzed the MCC across
all seven configurations, emphasizing model-level stability rather than
isolated feature performance. Qwen2.5-7B-Instruct demonstrated the
highest median MCC of 0.797~[0.67, 0.819], indicating superior overall
correlation between predictions and true labels, even under severe
imbalance, with consistent high scores for physiological features like
fever (0.8502) and seizures (0.8165). Llama-3.1-8B-Instruct demonstrated
a median MCC of 0.749~[0.67, 0.818], indicating strong and balanced
predictive performance. It achieved particularly high MCCs for symptoms such as pain (0.864) and respiratory complaints (0.842). However, slightly broader interquartile ranges were observed, reflecting moderate variability in features like doctor’s visit requests (0.6635). In contrast, smaller models such as Gemma-3-1B-it exhibited a lower median MCC of 0.502 [0.398, 0.581], indicating reduced overall reliability. Intermediate-sized models, including Llama-3.2-3B-Instruct and Qwen2.5-1.5B-Instruct, achieved median MCCs of 0.734 [0.656, 0.787] and 0.654 [0.589, 0.753], respectively, indicating a scale-dependent trend in which larger parameter counts are associated with improved overall concordance with ground truth labels. The multilingual Aya-expanse-8B variants demonstrated nuanced performance, with the English-translated input achieving a median MCC of 0.724 [0.634, 0.757], surpassing the Persian direct input at 0.686 [0.584, 0.757]. These results suggest that translation can enhance predictive correlation for specific classes without compromising overall model stability. Collectively, this metric affirms that mid-to-large SLMs maintain superior performance in resource-constrained and non-English environments.

\subsection{Larger models favored sensitivity while smaller models
achieved comparable or superior specificity}

\begin{figure}[!ht]
\centering
\includegraphics[width=0.8\textwidth]{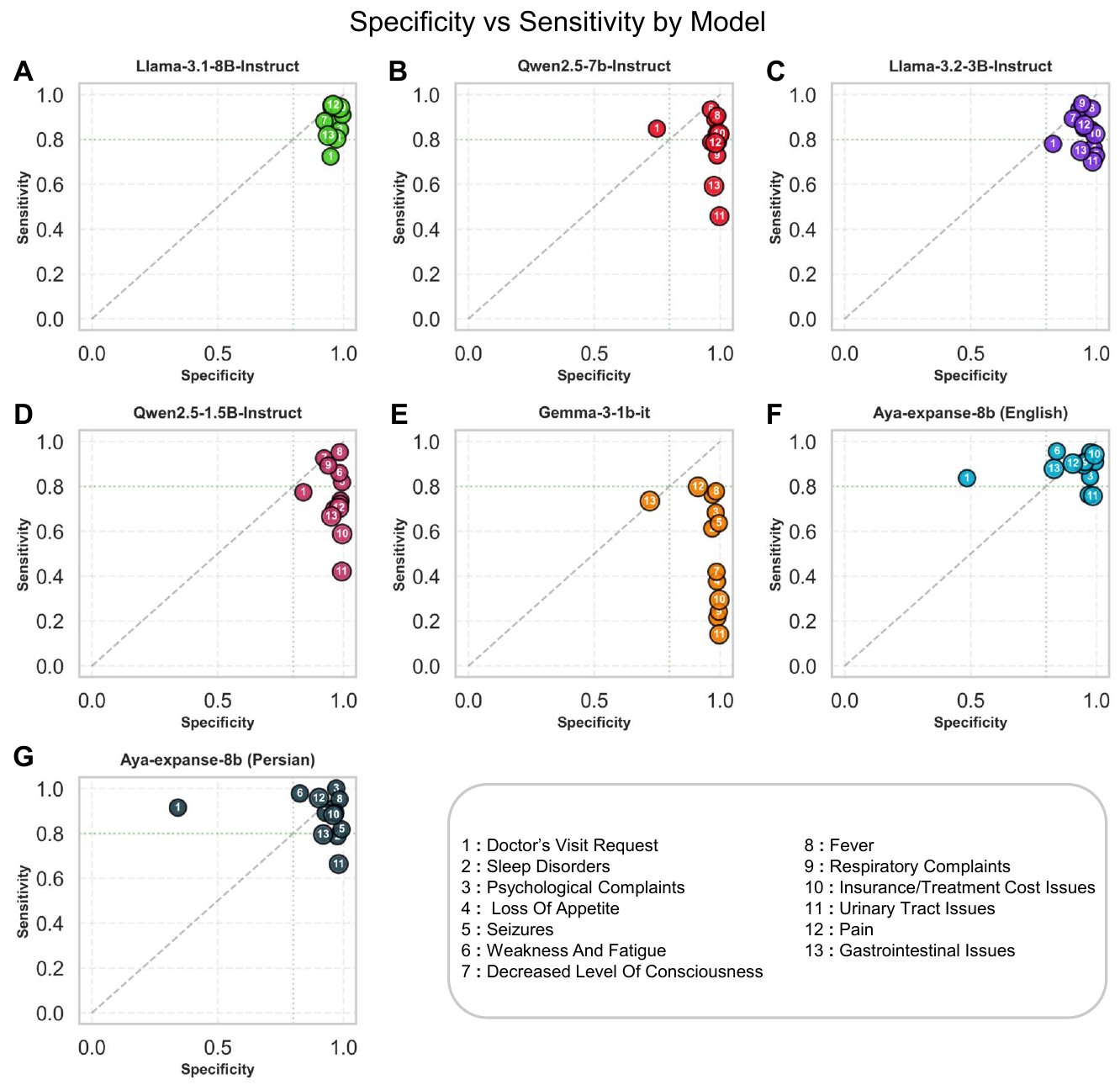} 
\caption{\textbf{Sensitivity–specificity trade-offs across evaluated small language models.} 
Each panel depicts sensitivity (y-axis) versus specificity (x-axis) for 13 binary clinical features (points labeled 1–13; see legend) for 
\textbf{(A)} Llama-3.1-8B-Instruct, 
\textbf{(B)} Qwen2.5-7B-Instruct, 
\textbf{(C)} Llama-3.2-3B-Instruct, 
\textbf{(D)} Qwen2.5-1.5B-Instruct, 
\textbf{(E)} Gemma-3-1B-it, 
\textbf{(F)} Aya-expanse-8B (English), and 
\textbf{(G)} Aya-expanse-8B (Persian). 
The dashed diagonal line denotes the locus where sensitivity equals specificity.}
\label{fig:3}
\end{figure}

To further evaluate model performance, sensitivity and specificity were computed for each of the 13 binary features, with corresponding correlations presented for each model in Figure ~\ref{fig:3}A–G. In terms of sensitivity, Llama-3.1-8B-Instruct emerged as the top performer with a median sensitivity of 0.909 [0.842–0.941], closely followed by Aya-expanse-8B (English) at 0.901 [0.842–0.911] and Aya-expanse-8B (Persian) at 0.893 [0.818–0.952]. Qwen2.5-7B-Instruct and Llama-3.2-3B-Instruct exhibited solid but more variable sensitivities of 0.818 [0.784–0.848] and 0.842 [0.762–0.893], respectively, with the former excelling in features such as insurance/treatment cost issues (0.824) but lagging on respiratory complaints (0.728). In contrast, smaller models exhibited markedly lower sensitivity. Qwen2.5-1.5B-Instruct achieved a median sensitivity of 0.737 [0.700–0.859], whereas Gemma-3-1B-it demonstrated the lowest sensitivity at 0.613 [0.294–0.735], with pronounced deficiencies in detecting features such as urinary tract issues (0.140) and doctor’s visit requests (0.215).

Regarding specificity, Qwen2.5-7B-Instruct achieved the highest median specificity of 0.987 [0.975–0.992], demonstrating exceptional precision in features such as seizures (0.998) and insurance issues (0.997), which is crucial for avoiding unnecessary escalations in resource-limited environments. Gemma-3-1B-it and Qwen2.5-1.5B-Instruct achieved median specificities of 0.986 [0.969–0.995] and 0.982 [0.951–0.988], respectively. Gemma-3-1B-it exhibited high specificity for psychological complaints (0.968), whereas Qwen2.5-1.5B-Instruct showed similarly high specificity for loss of appetite (0.987). Gemma-3-1B-it demonstrated particularly high specificity for psychological complaints (0.968), while Qwen2.5-1.5B-Instruct showed comparable specificity for loss of appetite (0.987). 

The larger Llama models also maintained high specificity, with Llama-3.1-8B-Instruct at 0.958 [0.951–0.982] and Llama-3.2-3B-Instruct at 0.951 [0.937–0.985], though slightly lower than the top-performing models and reduced for features such as gastrointestinal issues (0.938 and 0.937, respectively). The Aya-expanse-8B variants demonstrated balanced specificity, with the English version at 0.955 [0.906–0.976] and the Persian version at 0.960 [0.919–0.975]. Notably, the Persian input achieved higher specificity for pain (0.902) but showed reduced detection performance for features like doctor’s visit requests (0.342).

Overall, these model-wise differences indicate that increasing model size enhances sensitivity and supports more balanced coverage across features, while smaller models maintain comparable or even higher specificity.

\section{Discussion}

This study demonstrates that open-source SLMs, when
applied within a translation-augmented two-step pipeline, can effectively
extract structured clinical information from Persian palliative care
transcripts without the need for fine-tuning. Model performance varied
considerably, with larger instruction-tuned models demonstrating superior
MCC and sensitivity, particularly under class imbalance
conditions. Feature-wise, common physiological symptoms were extracted
reliably, whereas complex somatic, psychological, and administrative
features posed challenges across all models. Translating Persian
transcripts into English enhanced sensitivity and reduced missing outputs,
albeit with a modest decrease in specificity, highlighting a trade-off
with practical implications for pipeline design in low-resource language
contexts.

Building upon these findings, this study aligns with and extends several
concepts identified in previous works. Similar to the study by Akcali et
al.~\cite{ref23} on Turkish mammography reports, we demonstrate the
efficacy of prompt-engineered language models for clinical information extraction in
a low-resource language context. However, while their approach utilized a
massive, proprietary model (Gemini 1.5 Pro) in a many-shot setting, our
research pivots to evaluating smaller, open-source models in a few-shot
paradigm. This shift addresses the critical need for patient privacy and
accessible tools highlighted in the Navarro et al.\ study~\cite{ref24},
which noted a scarcity of freely available models and their rare
translation to clinical practice. Our work directly responds to this call
by benchmarking models that can run on desktop-grade hardware, similar to
the Strata framework described by Liu et al.~\cite{ref25}, which achieved
human-level performance with a fine-tuned Llama-3.1-8B model on a
curated dataset of pathology reports. Additionally, our analysis of
translation effects supports the findings of Balk et al.~\cite{ref26},
confirming that machine translation can serve as an effective
preprocessing strategy to reduce language bias. Specifically, we observed
enhanced sensitivity and data completeness when employing English prompts
with the multilingual Aya-expanse-8B model.

In comparison with recent methodological advances, our findings highlight the capabilities of general-purpose SLMs. Unlike the hybrid, multi-stage frameworks combining fine-tuned BERT with retrieval-augmented generation~\cite{ref27} or the sophisticated
multi-granularity embedding architectures proposed for psychomedical
named entity recognition (NER)~\cite{ref28}, our approach evaluates the
out-of-the-box few-shot capability of general-purpose SLMs. This
simplicity is a strategic advantage for rapid deployment.
Collectively, while specialized, finely-tuned architectures may achieve better accuracy on benchmark datasets, moderately sized, open-source SLMs with careful prompting present a compelling and immediately practical pathway for building privacy-preserving automated extraction systems in real-world, multilingual clinical settings with limited annotation resources and infrastructure. 

While this study demonstrates the potential of SLMs for clinical
information extraction in a low-resource language, several limitations
exist. These constraints include the relatively small scale of the
dataset and the inherent noise introduced by machine translation, which
may alter subtle semantic meanings. Also, our evaluation remains within a
single, specific domain of palliative care oncology, limiting
generalizability. Furthermore, the persistent underperformance on
administrative and psychological features suggests that current SLMs lack
the deep contextual and sociolinguistic understanding required for fully
reliable autonomous extraction. Building on these limitations, future
studies should prioritize several key directions, including using
large-scale, multi-center validation across diverse medical specialties
and languages to establish true generalizability and foster clinical
adoption.

In conclusion, this study establishes a meaningful proof of concept for
deploying open-source SLMs in privacy-preserving clinical NLP pipelines
beyond high-resource linguistic settings. By demonstrating that a
translation-augmented, few-shot prompting strategy enables structured
information extraction from Persian palliative oncology transcripts
without model fine-tuning, we offer a practical blueprint for healthcare
systems operating under infrastructure and data governance constraints.
The consistent advantage of 7B--8B parameter models in sensitivity and imbalance-aware performance metrics, combined with the trade-offs observed between native-language
and translated inference, underscores that a thoughtful combination of
model scale and language strategy is essential for deployment success.
Ultimately, integrating such systems into clinical workflows may reduce clinician documentation burden and expedite symptom-based triage for underserved populations. However, achieving this potential will require sustained interdisciplinary collaboration to develop linguistically diverse corpora, establish domain-specific evaluation frameworks, and implement robust human oversight mechanisms that satisfy the safety and ethical standards required in sensitive clinical settings.

\section{Methods}

This study aimed to evaluate and compare the performance of various SLMs
against a manually annotated dataset for the task of medical information
extraction. Utilizing a dataset of 1,221 transcribed phone calls from a
palliative care center, we benchmarked multiple open-source SLMs. The
process involved translating the original Persian transcripts, employing
a structured few-shot prompting strategy for feature extraction, and
conducting a comprehensive quantitative assessment using metrics tailored
for imbalanced binary classification. The following sections detail the
ethical approval, dataset construction, preprocessing pipeline, model
inference, and evaluation framework.

\subsection{Ethical consideration}

The study was approved by the Institutional Review Board (IRB) of
Isfahan University of Medical Sciences (IR.ARI.MUI.REC.1404.321). The
IRB granted this study a consent waiver. To protect patient
confidentiality, all data were pseudonymized before analysis. The study
adhered to the guidelines of the Helsinki Declaration (2013) and all
applicable regulations of the Iranian Ministry of Health.

\subsection{Data collection and dataset}

To establish a robust baseline for assessing SLM performance, we compiled a dataset comprising 1,221 transcribed phone calls to a cancer-focused palliative care call center. The transcripts, recorded in
Persian by operators, were anonymized by a single author (A.O.). They
were then divided into two groups, with two authors (S.G.\ and A.F.)
independently extracting medical information from each patient's reported
complaints. To ensure consistency, a blinded cross-review was conducted
between S.G.\ and A.F.\ All discrepancies were resolved by the
corresponding author (M.G.), ensuring a consistently annotated dataset
for comparative analysis. The extracted features are: doctor's visit
request, psychological complaints, sleep disorders, loss of appetite,
seizures, weakness and fatigue, decreased level of consciousness, fever,
respiratory complaints, insurance/treatment cost issues, urinary tract
issues, pain, and gastrointestinal issues. Collectively, they are
essential for assessing patient needs and guiding resource allocation.

\subsection{Data preprocessing}

To preprocess our dataset of 1,221 phone calls in Persian, we first used
a translation model to convert each sample into English. Then, we applied
a few-shot setting to each sample to generate the final input prompts for
five different models with varying numbers of parameters.

We used the Aya-expanse-8B model, a multilingual SLM supporting 23
languages, including Persian, as our translator. The generation
configuration was set with a temperature of 0.3 and a maximum of 2048
new tokens. For each sample in our dataset, we prompted the model with
the instruction ``Translate the following text from Persian to English:''
to obtain its English translation.

To prepare the textual data for structured information extraction, we employed a few-shot prompting strategy. This approach involved providing a clear system prompt alongside three examples of input and output data from the dataset to guide the model in generating outputs in a predefined format. The full input prompt, including few-shot examples, is presented in Supplementary File 2.

The system prompt defined the model's role and constraints as follows:

\begin{quote}
\textit{You are an expert in data extraction specializing in medical
information. You are provided with clinical data about patients with
cancer who require palliative care. Your task is to read the patient's
condition and extract ONLY complications strictly into the predefined
format below.}

\textit{Specific instructions were included to ensure consistency:}
\begin{enumerate}
  \item \textit{Output exactly the same structure and order of fields.}
  \item \textit{Fill each field with True or False only.}
  \item \textit{Do not infer or assume any information that is not
        explicitly stated.}
  \item \textit{Do not output any text outside the provided template.}
\end{enumerate}
\end{quote}

The output template comprised 13 binary fields covering a range of
potential complications and concerns, including: patient-requested
visits, psychiatric complaints, sleep disorders, appetite loss, seizures,
weakness/fatigue, decreased consciousness, fever, respiratory complaints,
insurance/cost issues, urinary tract issues, pain, and gastrointestinal
issues. In addition, for extra evaluation of the translator model, we
used an identical system prompt in Persian for processing the original
transcripts.

\subsection{Generation and post-processing}

We evaluated five SLMs with varying parameter counts for text generation.
The models included: Qwen2.5-7B-Instruct, Llama-3.1-8B-Instruct,
Llama-3.2-3B-Instruct, Gemma-3-1B-it, and Qwen2.5-1.5B-Instruct,
alongside Aya-expanse-8B as the translator model. All experiments were
conducted using HuggingFace’s Transformers library (version 4.57.3) ~\cite{ref29} with sampling
disabled and a limit of 512 new tokens per generation. Following the
generation phase, a custom function was employed to detect and extract
each precise feature from the model outputs. The resulting feature sets
were subsequently structured into tables to enable comparison with the
manually extracted ground truth data. Additionally, we quantified missing
predictions---cases where models failed to output a valid binary
value---and treated these as negative (False) instances in final
analyses, reflecting the model's inability to detect the feature.

\subsection{Assessment}

To evaluate each model's performance, we employed a suite of
complementary metrics capturing global accuracy, class-wise behavior,
and robustness to class imbalance. Specifically, we computed accuracy,
specificity, sensitivity, precision, the macro-averaged F1
score, and MCC.

Accuracy was used to summarize the overall proportion of correct
predictions. Sensitivity and precision were analyzed to assess the models’ ability to detect positive cases. Sensitivity measured the proportion of true positives correctly identified, reflecting the model’s capacity to avoid false negatives, while precision quantified the proportion of predicted positives that were correct, indicating how well the model minimizes false positive predictions.
Specificity quantified the correct rejection of negative cases, which is
essential in imbalanced settings. To provide a class-balanced measure of
predictive quality, we computed the macro-averaged F1 score, preventing
inflation from majority classes.

To quantify global agreement between predicted and true labels, we used
the MCC, defined as:

\begin{equation}
\text{MCC} = \frac{TP \times TN - FP \times FN}
             {\sqrt{(TP + FP)(TP + FN)(TN + FP)(TN + FN)}}
\end{equation}

which provides a stable summary even under severe class imbalance.

All statistical analyses and metric computations were performed in Python
(version 3.13.5) ~\cite{ref30} using pandas (version 2.3.3) ~\cite{ref31}, NumPy (version 2.3.5) ~\cite{ref32},
SciPy (version 1.16.3) ~\cite{ref33}, and scikit-learn (version 1.7.2) ~\cite{ref34}.

\subsection{Computational environment}

All inferences for the five SLMs and the translator model were performed
locally using a single L4 GPU (24 GB VRAM) and a two-core CPU with 8 GB
of RAM, with no external API calls, which preserves privacy.

\newpage
\section*{Data availability}

The data generated and/or utilized in this study can be obtained from the
corresponding author upon reasonable request
(\href{mailto:mreghafarzadeh@gmail.com}{mreghafarzadeh@gmail.com}).

\section*{Code availability}

The code for all experiments and analyses is publicly available at
\url{https://github.com/mohammad-gh009/Small-language-models-on-clinical-data-extraction.git}.

\section*{Acknowledgments}

The authors would like to express their sincere gratitude to Ali Motahharynia for providing insightful comments that greatly improved the quality of this paper.

\section*{Funding}

No funding was received for this study or its publication.

\section*{Competing interests}

The authors declare no competing interests.

\section*{Author contributions}

Conceptualization: M.G, A.B. Supervision: M.G. Dataset preparation: M.G,
A.O, S.G, A.F. Model development: M.G. Model assessment: M.G, N.Y, E.H.
Data interpretation: All authors. Drafting original manuscript: M.G, N.Y,
E.H. Revising the manuscript: M.G, A.B. All the authors have read and
approved the final version for publication and agreed to be responsible
for the integrity of the study.


\end{document}